\documentclass[10pt,twocolumn,letterpaper]{article}

 \usepackage{iccv}              

\usepackage{tikz}
\usetikzlibrary{positioning, fit, arrows}

\definecolor{iccvblue}{rgb}{0.21,0.49,0.74}
\usepackage[pagebackref,breaklinks,colorlinks,allcolors=iccvblue]{hyperref}

\def\paperID{4} 
\def\confName{ICCV}
\def\confYear{2025}

\title{Pseudo-Label Refinement for Robust Wheat Head Segmentation via Two-Stage Hybrid Training}

\author{
Jiahao Jiang\thanks{Equal contribution to this work.}\quad
Zhangrui Yang\footnotemark[1]\quad
Xuanhan Wang \quad
Jingkuan Song \\
Tongji University\\
{\tt\small zaleni233@outlook.com, jerry.zryang@hotmail.com, wxuanhan@hotmail.com, jingkuan.song@gmail.com}
}

\begin{document}
\maketitle 
\thispagestyle{empty}
\pagestyle{empty}
\section{Introduction}
\label{sec:intro}

Accurate semantic segmentation of wheat heads is crucial for various agricultural applications, which involves many challenges~\cite{najafian2023semi,ghanbari2024semi} such as limited labeled data and diverse field conditions. This significantly limits the generalization of fully supervised models and requires effective strategies that can leverage all available data, especially from those extensive unlabeled data.

Inspired by prior works~\cite{10704987,wu2021optimized,cheng2022masked,xie2021segformer,kirillov2023segment,ronneberger2015u,long2015fully} for fine-grained segmentation, we develop a self-supervised training framework that iteratively refines pseudo-labels through a teacher–student loop. In particular, we propose a two-stage hybrid training strategy. This effectively combines pseudo-label pre-training with high-resolution fine-tuning on ground-truth data. This dual approach ensures both broad feature learning and fine detail capture.

\section{Approach}
\label{sec:formatting}

\begin{figure}[h]
    \centering
    \resizebox{0.75\linewidth}{!}{  
    \begin{tikzpicture}[
        block/.style={rectangle, draw, fill=teal!60, rounded corners,
            text width=6.5em, text centered, minimum height=2em, node distance=0.5cm, font=\footnotesize, text=white},
        dashed_block/.style={rectangle, draw, dashed, inner sep=8pt, line width=0.8pt},
        arrow/.style={thick,->,>=stealth},
        feedback_arrow/.style={thick,->,>=stealth, bend left=90},
        label_node/.style={align=center, font=\footnotesize}
    ]
    \node (stage0) [block] {Stage0: Start Model Training};
    \node (generate) [block, below=of stage0] {Generate: Pseudolabel Generation};
    \node (stage1) [block, below=of generate] {Stage1: Pseudolabel Data (More) Train from Scratch};
    \node (stage2) [block, below=of stage1] {Stage2: Real-label Data (Less) Fine-tune};
    \node (eval) [block, below=of stage2, fill=none, draw=none, text=black] {Evaluate};

    \draw [arrow] (stage0) -- (generate);
    \draw [arrow] (generate) -- (stage1);
    \draw [arrow] (stage1) -- (stage2);
    \draw [arrow] (stage2) -- (eval);
    \draw [arrow] (stage2.west) -- ++(-1.2,0) |- (generate.west);

    \node [label_node, left=of stage2.west, xshift=-0.5cm, yshift=1.2cm] (teacher_model) {As Teacher Model};
    \node [dashed_block, fit=(stage1) (stage2), label={[xshift=0.6cm, yshift=-0.8cm, rotate=90, font=\footnotesize]right:Student Model}] (student_model_box) {};
    \end{tikzpicture}
    }
    \caption{Overview of our self-training method. The process begins with a base model, which generates pseudo-labels for unlabeled data. The student model is then trained in two stages: first on pseudo-labeled data, and then fine-tuned on true-labeled data.}
    \label{fig:method}
\end{figure}
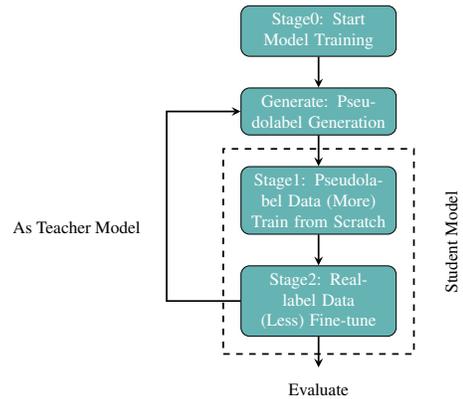

Our methodology is built upon a self-training mechanism, drawing inspiration from knowledge distillation concepts\cite{hinton2015distilling}. The core of our approach utilizes the SegFormer model \cite{xie2021segformer}, specifically with a Mix Transformer (MiT-B4) as its backbone. We use the same architecture—SegFormer with a MiT-B4 backbone—in all training stages. Each stage shares this structure. Only the training data and hyperparameters differ. This setup maintains consistency while allowing task-specific adaptation. The overall architecture and strategy are illustrated in Figure \ref{fig:method}.

The process begins with ``Stage 0: Start Model Training''. We first train a model using only the limited labeled dataset. This model serves as the initial “teacher.” It generates pseudo-labels for the unlabeled data. These labels form the basis for the next training stage. The ``student'' model then undergoes a two-stage training. In ``Stage 1: Pseudolabel Data (More) Train from Scratch,'' the student model is trained from scratch on a larger set of pseudo-labeled data. Subsequently, in ``Stage 2: Real-label Data (Less) Fine-tune,'' the student is fine-tuned using a smaller set of true-labeled data. The refined student model's performance is evaluated. A critical aspect of our approach is the iterative feedback loop: the refined student model from Stage 2 can itself become a new ``teacher'' model, generating improved pseudo-labels for subsequent training iterations, thus enabling a self-improving training loop.

\subsection{Dataset Utilization}

Our dataset strategy involved a phased expansion for robust training. We began by fine-tuning a model pre-trained on ImageNet-1k to generate pseudo-labels using the provided pre-train dataset. These pseudo-labels were meticulously filtered based on a set confidence threshold to ensure high quality for our training set. The provided 99 masked training data points were then used for further fine-tuning, progressively refining performance. This iterative process was repeated multiple times, leading to a gradual improvement in pseudo-label quality.

Data augmentation played a crucial role in preventing overfitting and enhancing model generalization. Our augmentation pipeline included: random cropping, horizontal and vertical flipping, 90-degree rotations, scaling ($\pm$20\%), translation ($\pm$6.25\%), rotation ($\pm$30$^\circ$), brightness/contrast adjustments ($\pm$30\%), HSV transformations, and ImageNet normalization.

\subsection{Novel Contributions}

Our primary contribution is a systematic self-training framework specifically tailored for the wheat segmentation task. This framework optimizes data utilization and progressively enhances model accuracy through two key innovations:

\begin{enumerate}[leftmargin=*, itemsep=0.2em]
    \item \textbf{Iterative Teacher-Student Loop:} We employ a cyclical learning process where a refined ``student'' model transforms into the ``teacher'' for the next iteration. This mechanism continuously generates higher-quality pseudo-labels from unlabeled data, establishing a self-improving loop that learns from the entire dataset.
    \item \textbf{Two-Stage Hybrid Training Strategy:} Within each cycle, we introduce a specialized two-stage training regimen. The model is initially pre-trained on a large set of high-confidence pseudo-labels at a lower resolution (512x512) to learn robust general features. Subsequently, it is fine-tuned on the high-resolution (1024x1024) ground-truth data to capture fine details.
\end{enumerate}
\section{Experiments}
\subsection{Experimental Settings}

Training was conducted on a single NVIDIA RTX 4090 D 24GB GPU (CUDA 12.4) running Ubuntu 20.04.6 LTS. We utilized a two-stage training strategy, with the second stage incorporating 10-fold cross-validation.
\begin{itemize}[leftmargin=*, itemsep=0.1em]
    \item \textbf{Stage 1 (Pseudo-label Pre-training):} The model was trained for 40 epochs on 512x512 images with a batch size of 8 and a learning rate of $6 \times 10^{-5}$.
    \item \textbf{Stage 2 (Fine-tuning):} The model was fine-tuned for 25 epochs per fold on 1024x1024 images with a batch size of 1. We employed the AdamW optimizer with an initial learning rate of $1 \times 10^{-5}$.
\end{itemize}
Data augmentation served as our primary regularization technique throughout the training process.

\subsection{Inference \& Post-processing}

Our inference pipeline is designed for high accuracy and robustness. We implement a model ensembling strategy by averaging predictions from multiple well-performing models trained during the 10-fold cross-validation. To further enhance performance, extensive Test-Time Augmentation (TTA) is applied to each model. This includes processing the original image, horizontal and vertical flips, 90-degree rotations, and multi-scale inputs (0.75x, 1.25x). The logits from each augmented view are transformed back to their original orientation and then averaged. These TTA-enhanced predictions are subsequently averaged across all ensembled models. The final segmentation mask is obtained by upsampling the ensembled logits to the input resolution (1024x1024), performing an argmax operation, and then resizing the result to the required 512x512 output format using nearest-neighbor interpolation before saving as a CSV file.

\subsection{Results}

Our model's performance was rigorously evaluated on both the Development Phase and Testing Phase datasets, using the competition-provided metrics. The key results are summarized in Table \ref{tab:results}.

\begin{table}[h]
\centering
\begin{tabular}{lcc}
\toprule
\textbf{Dataset} & \textbf{mIoU} \\
\midrule
Development Phase & 0.7480 \\
Testing Phase & 0.7099 \\
\bottomrule
\end{tabular}
\caption{Performance results on competition datasets}
\label{tab:results}
\end{table}

\begin{figure}[t]
\centering
\includegraphics[width=0.75\linewidth]{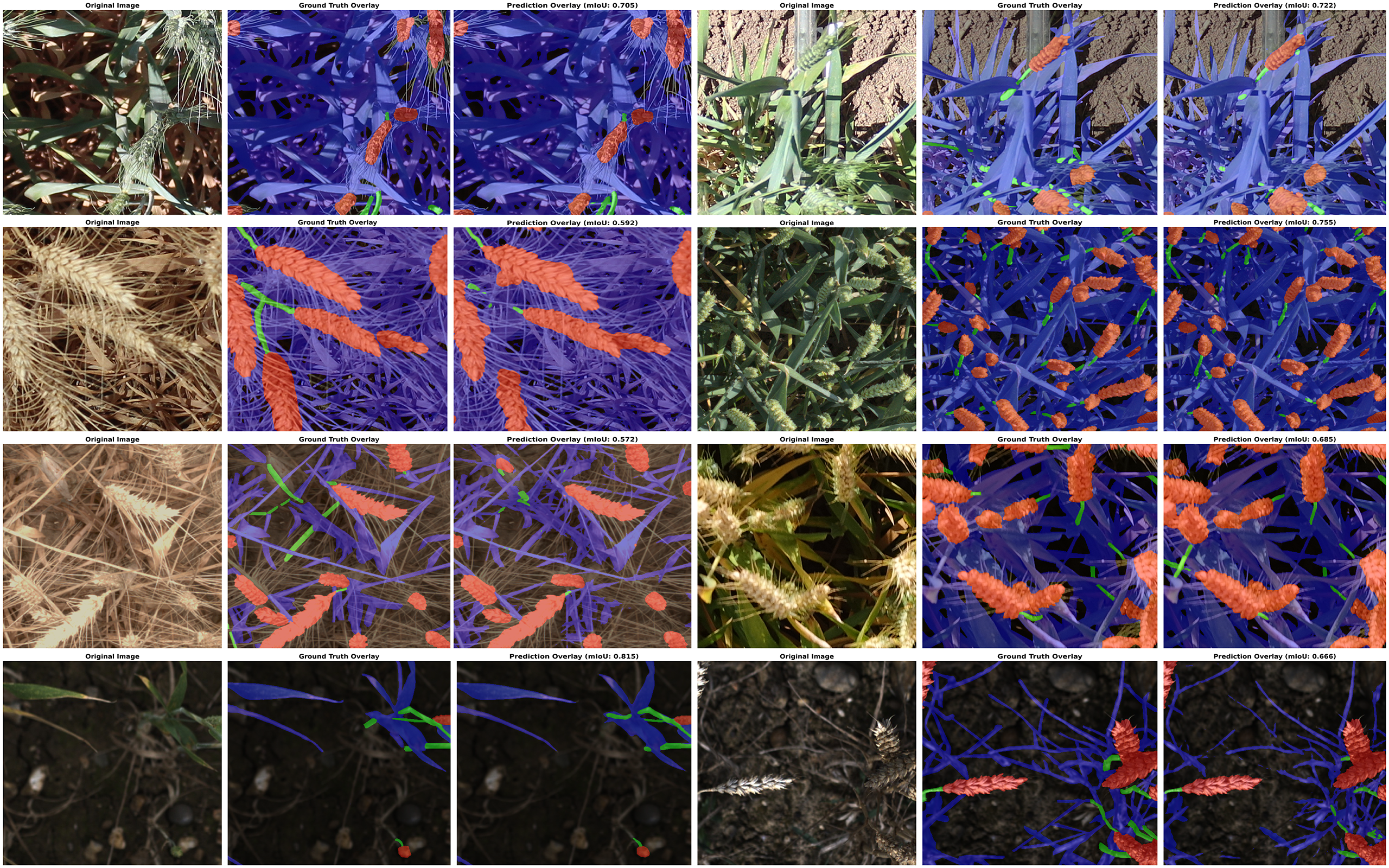}  
\caption{
Qualitative results on eight random test samples, including input images, ground-truth, and predictions
}
\label{fig:segmentation_results}
\end{figure}

Figure \ref{fig:segmentation_results} shows that the model effectively segments wheat heads, demonstrating robustness across challenging scenarios.

\section{Conclusion}

We presented a comprehensive solution for the Global Wheat Full Semantic Segmentation Competition, leveraging a self-training framework with an iterative teacher-student loop and a two-stage hybrid training strategy. Our approach effectively utilized both labeled and unlabeled data through high-confidence pseudo-labeling and progressive refinement. The integration of SegFormer as the backbone, coupled with extensive data augmentation and an advanced inference pipeline incorporating model ensembling and Test-Time Augmentation, contributed to our competitive results. Future work could explore more investigate the impact of different backbone architectures within this iterative framework.

{
    \small
    \bibliographystyle{ieeenat_fullname}
    \bibliography{main}
}

\end{document}